\documentclass[letterpaper]{article} 
\usepackage[submission]{aaai2026}  
\usepackage{times}  
\usepackage{helvet}  
\usepackage{courier}  
\usepackage[hyphens]{url}  
\usepackage{graphicx} 
\usepackage{natbib}  
\usepackage{caption} 
\usepackage{algorithm}
\usepackage{algorithmic}
\usepackage{amsmath}
\usepackage{amsfonts}       
\usepackage{amssymb}        
\usepackage{booktabs}       
\usepackage{multirow}
\usepackage{authblk}

\usepackage{newfloat}
\usepackage{listings}
\DeclareCaptionStyle{ruled}{labelfont=normalfont,labelsep=colon,strut=off} 
\floatstyle{ruled}
\newfloat{listing}{tb}{lst}{}
\floatname{listing}{Listing}
\title{Constraints-Guided Diffusion Reasoner for Neuro-Symbolic Learning}

\author[1,2]{Xuan Zhang}
\author[1,2]{Zhijian Zhou}
\author[4]{Weidi Xu}
\author[5]{Yanting Miao}
\author[4]{Chao Qu\textsuperscript{\textdagger}} 
\author[1,3]{Yuan Qi\textsuperscript{\textdagger}} 

\affil[1]{Fudan University}
\affil[2]{Shanghai Innovation Institute}
\affil[3]{Shanghai Academy of Artificial Intelligence for Science}
\affil[4]{INFTECH}
\affil[5]{University of Waterloo}

\usepackage{bibentry}

\begin{document}

\maketitle
\begingroup
\renewcommand\thefootnote{\textdagger} 
\footnotetext{Corresponding authors.}
\addtocounter{footnote}{-1} 
\endgroup

\begin{abstract}
Enabling neural networks to learn complex logical constraints and fulfill symbolic reasoning is a critical challenge. Bridging this gap often requires guiding the neural network’s output distribution to move closer to the symbolic constraints. While diffusion models have shown remarkable generative capability across various domains, we employ the powerful architecture to perform neuro-symbolic learning and solve logical puzzles. Our diffusion-based pipeline adopts a two-stage training strategy: the first stage focuses on cultivating basic reasoning abilities, while the second emphasizes systematic learning of logical constraints. To impose hard constraints on neural outputs in the second stage, we formulate the diffusion reasoner as a Markov decision process and innovatively fine-tune it with an improved proximal policy optimization algorithm. We utilize a rule-based reward signal derived from the logical consistency of neural outputs and adopt a flexible strategy to optimize the diffusion reasoner's policy. We evaluate our methodology on some classical symbolic reasoning benchmarks, including Sudoku, Maze, pathfinding and preference learning. Experimental results demonstrate that our approach achieves outstanding accuracy and logical consistency among neural networks. 
\end{abstract}

\begin{links}
    \link{Code}{https://github.com/dd88s87/DDReasoner}
\end{links}


\section{Introduction}
Despite the rapid advance of deep learning, it is still challenging for deep neural networks to solve problems with complex and hard constraints such as symbolic reasoning. For instance, the output of a neural network may achieve good approximation over most regions; however, local prediction errors can still result in outputs that violate the constraints, thus failing to obtain a correct solution. To address this issue, there has been an increasing amount of neuro-symbolic learning methods to inject domain knowledge constraints and improve the performance of complex reasoning. These methods primarily involve modifying the loss function \cite{pmlr-v80-xu18h, Wang2019IntegratingDL, NEURIPS2019_7b66b4fd, li2019consistency, ahmed2023a}, incorporating constraint layers \cite{giunchiglia2021, NEURIPS2022_c182ec59}, and integrating logic engines or algorithmic components \cite{ DBLP:journals/corr/abs-1905-12149, cornelio_2023_NASR, li2023neurosymbolic, agarwal2021endtoendneurosymbolicarchitectureimagetoimage, ijcai2020p243, petersen2021learning}, among others. However, these methods may suffer from poor performance or expensive time consumption. 

Parallel to this, diffusion models \cite{NEURIPS2020_4c5bcfec, Sohl-Dickstein_Weiss_Maheswaranathan_Ganguli_2015, Nichol_Dhariwal_2021} have emerged as an outstanding paradigm of generative models and demonstrated immense potential in many related downstream tasks \cite{saharia2022photorealistictexttoimagediffusionmodels, dhariwal2021diffusion, rombach2022highresolutionimagesynthesislatent, ho2022video, he2023latentvideodiffusionmodels, janner2022diffuser, chi2023diffusionpolicy}. By learning how to denoise from noisy and corrupted samples, diffusion models can generate high-quality samples approximating the data distribution. Due to the inherent iterative and stochastic characteristics, some recent studies apply diffusion architectures to reasoning or planning tasks \cite{janner2022diffuser, chi2023diffusionpolicy, ye2024diffusion, ye2025beyond, zhao2025d1scalingreasoningdiffusion, Du_2024_ICML, zhang2025tscendtesttimescalablemctsenhanced, fan2023reinforcement, black2024trainingdiffusionmodelsreinforcement}, which are primarily combined with text, visual information, or robot trajectories.

In this work, we apply denoising diffusion probabilistic models (DDPMs) to perform symbolic reasoning, i.e., \textbf{DDReasoner}. We employ a two-stage pipeline to train our neural network: (1) Basic Supervised Learning (SL): Other related efforts focused on boosting the performance of diffusion models in this stage \cite{Du_2024_ICML, 10.1145/3637528.3671783}. In this work, we use the simplest form of masked DDPM training to enable DDReasoner to preliminarily possess the capability to solve logical puzzles; (2) A Novel Application of Reinforcement Learning (RL): Generated solutions with minor errors in a few positions, which lead to the violation of overall constraints, may exhibit a minimal loss value during the SL phase, indicating that DDReasoner may fail to identify them as suboptimal outputs. Therefore, we introduce a novel application of RL. By imposing hard logical constraints on the reward function, we further guide DDReasoner to internalize constraints and refine its reasoning trajectory. During RL training, following the method of \citet{black2024trainingdiffusionmodelsreinforcement}, we formulate the iterative denoising process as a Markov Decision Process (MDP), in which the termination state represents DDReasoner’s predicted solution. We utilize a rule-based reward function derived from the logical consistency of the generated solution and employ a dynamic and efficient fine-tuning strategy to optimize the policy. Despite previous explorations about logically-constrained RL \cite{fu2014probablyapproximatelycorrectmdp, hasanbeig2019logicallyconstrainedreinforcementlearning}, we combine it with the representation capabilities of deep neural networks and develops this paradigm that operates effectively in low-dimensional discrete spaces while also offering transferability to high-dimensional spaces. To the best of our knowledge, our work is the first to utilize RL techniques for training diffusion models to solve symbolic logical reasoning tasks.

We evaluate our methodology on some symbolic reasoning benchmarks, including Sudoku, Maze, simple path prediction \& preference learning and Minimum-cost path finding. Experimental results indicate that our integration of RL leads to a significant improvement in the performance of DDReasoner trained using supervised learning, and the final results also outperform purely neural network-based methods with other architectures. Experiments on Maze also reveal that our RL method unlocks the full potential of a foundational model, enabling it to achieve perfect accuracy while maintaining high efficiency regarding the parameter count and data requirements. Furthermore, our cross-dataset and varying-size experiments, detailed in Appendix B.4, confirm the superior generalizability of our RL approach.

Our contributions can be summarized as follows: (1) We innovatively employ reinforcement learning in our training pipeline, enabling diffusion models to internalize hard logical constraints; (2) We evaluate our methodology on some symbolic reasoning benchmarks. The outstanding performance demonstrates that our constraints-guided RL method facilitates the application of diffusion models to symbolic reasoning tasks that are based on constraint satisfaction.

\section{Related Work and Motivation}
\label{related_work}
\subsection{Neuro-Symbolic Learning}
Despite the rapid advancement of deep learning, it remains challenging for deep neural networks to solve certain classes of problems, particularly those requiring the ability to apply logical rules to analyze information, identify connections, and make consistent decisions, i.e., symbolic reasoning. 

There has been a lot of recent research regarding the imposition of constraints in neural networks to bridge the gap between deep neural networks and logical constraints, i.e., neuro-symbolic learning. A common class of work is modifying the loss function to quantify the level of disagreement with logical constraints and encourage the neural output to better satisfy the given constraints \cite{pmlr-v80-xu18h, Wang2019IntegratingDL, NEURIPS2019_7b66b4fd, li2019consistency, ahmed2023a}. Some work added a layer to the neural network to manipulate the output closer to the logical constraints \cite{giunchiglia2021, NEURIPS2022_c182ec59}. However, these supervised learning-based methods frequently lead to challenges in generalization to adhere to constraints across unseen problems. Some other approaches attempt to integrate symbolic solvers or algorithmic components, which guarantees nearly perfect correctness, into deep learning architectures \cite{DBLP:journals/corr/abs-1905-12149, cornelio_2023_NASR, li2023neurosymbolic, agarwal2021endtoendneurosymbolicarchitectureimagetoimage, ijcai2020p243, petersen2021learning}. In particular, \citet{DBLP:journals/corr/abs-1905-12149} introduced a differentiable maximum satisfiability (MAXSAT) solver into a traditional convolutional architecture. \citet{cornelio_2023_NASR} constructed a pipeline consisting of a neural solver, a hard-attention mask predictor and a symbolic solver. 

Although integrating with a symbolic solver or algorithm yielded favorable results on corresponding tasks, they do not enable neural networks to truly acquire the capability to solve such problems independently. In our work, we aim to apply the diffusion model as a reasoner to solve complex symbolic reasoning tasks that require strict constraints. We utilize task-specific symbolic solvers or computational algorithms only once during the initial ground truth generation. Throughout the training process, these solvers or algorithms do not need to be re-executed; At inference time, our diffusion reasoner maintains a commendable level of accuracy and consistency without the need for auxiliary modules.

\begin{figure*}
  \centering
  \includegraphics[width=0.7\linewidth]{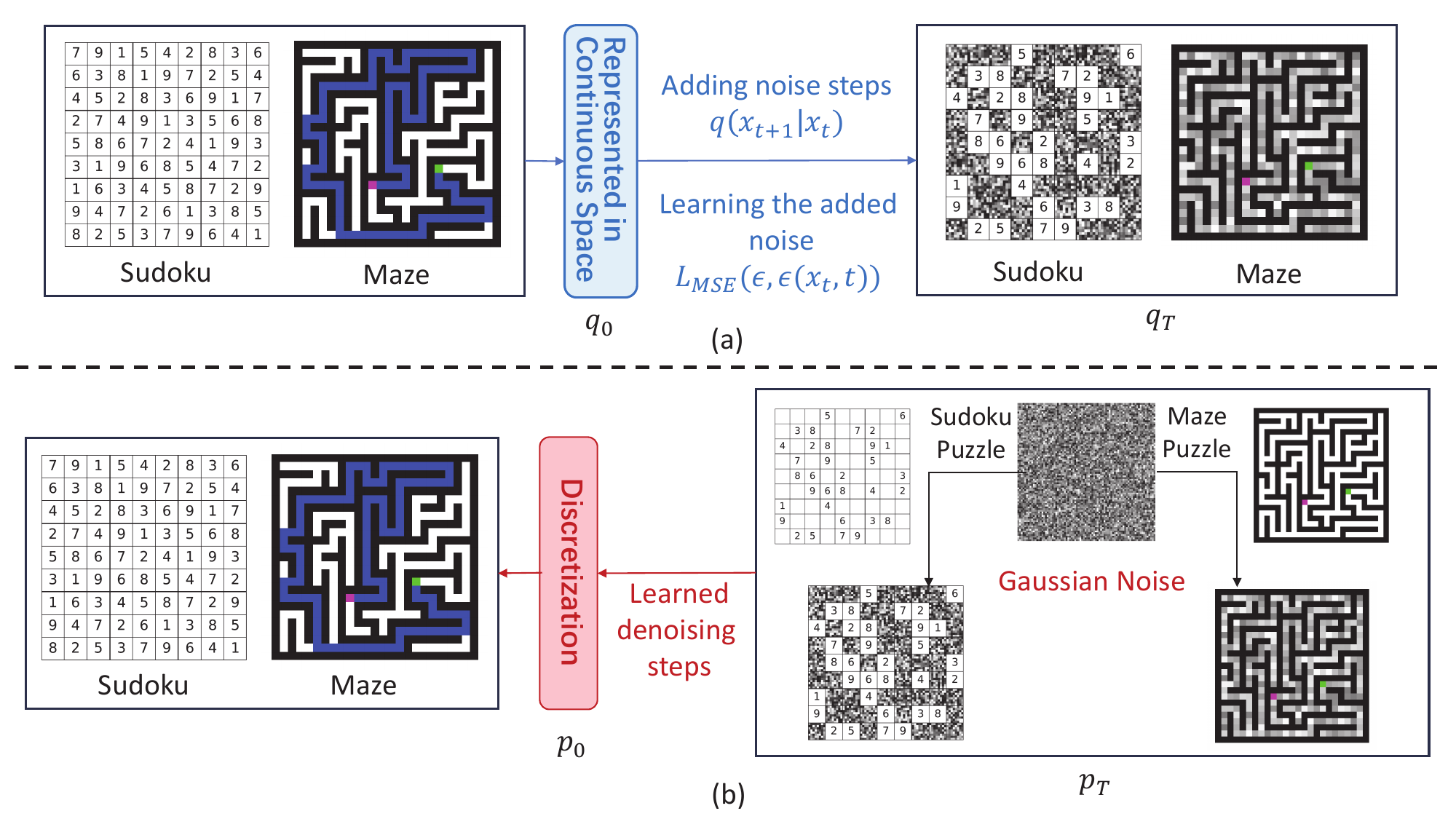}
  \caption{Overview: examples of DDReasoner solving Sudoku and Maze puzzles. (a) Supervised learning. (b) At inference time (greedy deterministic sampling). $q_0, ...,q_T$ and $p_T, ..., p_0$ are all representations in the continuous space, but for illustrative purposes, we render the digits at each cell (Sudoku) and walls (Maze) in $q_T$ and $p_T$ in a discrete format—representing the discretizations of distributions in the continuous space—while the regions with noised distributions are visualized as noise.}
  \label{diffusion_pic}
\end{figure*}
\subsection{Diffusion Models as Sequential Decision-Makers}
Diffusion models \cite{Sohl-Dickstein_Weiss_Maheswaranathan_Ganguli_2015, NEURIPS2020_4c5bcfec, Nichol_Dhariwal_2021} have emerged as important generative models for images \cite{saharia2022photorealistictexttoimagediffusionmodels, dhariwal2021diffusion}, videos \cite{ho2022video, he2023latentvideodiffusionmodels}, robotic control \cite{janner2022diffuser, chi2023diffusionpolicy}, and many other modalities. These models work by gradually adding noise to a data distribution in a forward process until it becomes a simple distribution, then learning to reverse the process step by step to recover original data, i.e., denoising. 

Due to the inherent property of iterative refinement, the denoising process can be regarded as a sequential decision-making problem. Recent approaches conceptualized a diffusion model as a reasoner or planner, especially in the domain of natural language reasoning \cite{ye2024diffusion, ye2025beyond, zhao2025d1scalingreasoningdiffusion}, robotic control \cite{janner2022diffuser, chi2023diffusionpolicy} and visual generation \cite{fan2023reinforcement, 
black2024trainingdiffusionmodelsreinforcement, fan2024optimizingddpmsamplingshortcut, wallace2023diffusionmodelalignmentusing, gupta2025simpleeffectivereinforcementlearning}. In the field of visual generation, some work has emerged that employs reinforcement learning to fine-tune image synthesis processes. \citet{fan2024optimizingddpmsamplingshortcut} introduced a policy gradient-based algorithm to optimize the generated distribution, allowing the DDPM sampler to explore more efficient sampling trajectories. \citet{fan2023reinforcement} treated the denoising process as a Markov Decision Process (MDP) and proposed DPOK, a KL-regularized policy gradient algorithm to align text-to-image diffusion models with human preferences.  \citet{black2024trainingdiffusionmodelsreinforcement} proposed DDPO and applied Proximal Policy Optimization (PPO) \cite{schulman2017proximalpolicyoptimizationalgorithms} to update the policy, while \citet{wallace2023diffusionmodelalignmentusing} adapted Direct Preference Optimization (DPO) \cite{rafailov2023direct} to diffusion models. More recently, \citet{gupta2025simpleeffectivereinforcementlearning} proposed LOOP, integrating the respective advantages of PPO and RLOO \cite{ahmadian-etal-2024-back}.

While some existing studies have made attempts to employ continuous or discrete diffusion models to perform complex symbolic reasoning and planning \cite{Du_2024_ICML, zhang2025tscendtesttimescalablemctsenhanced, 10.1145/3637528.3671783, vankrieken2025neurosymbolicdiffusionmodels, suresh2025dingo}, the integration of reinforcement learning techniques remains unexplored. To analyze the effectiveness of RL methods in addressing such tasks, we adopt a flexible and efficient policy optimization 
algorithm to fine-tune the diffusion reasoner, enabling it to internalize logical constraints and enhancing its ability to generate compliant solutions at inference time.

\section{Preliminaries}
\subsection{Diffusion Models}
\label{diffusion}
The generative process of a denoising diffusion probabilistic model (DDPM) \cite{NEURIPS2020_4c5bcfec} relies on the sampler $p_\theta(\textbf{x}_{t-1} | \textbf{x}_t)$ to reverse the forward process. It is parameterized by a neural network $\epsilon_\theta(\textbf{x}_t, t)$, which predicts the added noise $\tilde{\epsilon}(\mathbf{x}_0, t)$ that transforms $\textbf{x}_0$ to $\textbf{x}_T$ with the following objective:
\begin{align}
    \mathcal{L}_{\text{DDPM}}(\theta) = \mathbb{E}_{\mathbf{x}_0, t , \mathbf{x}_t} \left[ \left\| \tilde{\epsilon}(\mathbf{x}_0, t) - \epsilon_\theta(\mathbf{x}_t, t) \right\|^2 \right]
\end{align}

Sampling starts from a pure Gaussian noise $p(\textbf{x})$ by iteratively denoising following $p_\theta(\textbf{x}_{t-1} | \textbf{x}_t)$ to produce a trajectory $\{ \textbf{x}_T, \textbf{x}_{T-1}, ..., \textbf{x}_0\}$. The sampler at each step is expressed as:
\begin{align}
    p_\theta(\textbf{x}_{t-1}|\textbf{x}_t) = \mathcal{N}(\textbf{x}_{t-1}| \mu_t(\textbf{x}_t, \epsilon_\theta(\textbf{x}_t, t)), \sigma_t^2\textbf{I})
\end{align}

Above, $\mu_t$ is a function that maps $\textbf{x}_t$ and $\epsilon_\theta$ to the mean timestep $t-1$, and $\sigma_t^2$ is the variance term. 

\label{method}
\begin{figure*}
  \centering
  \includegraphics[width=0.8\linewidth]{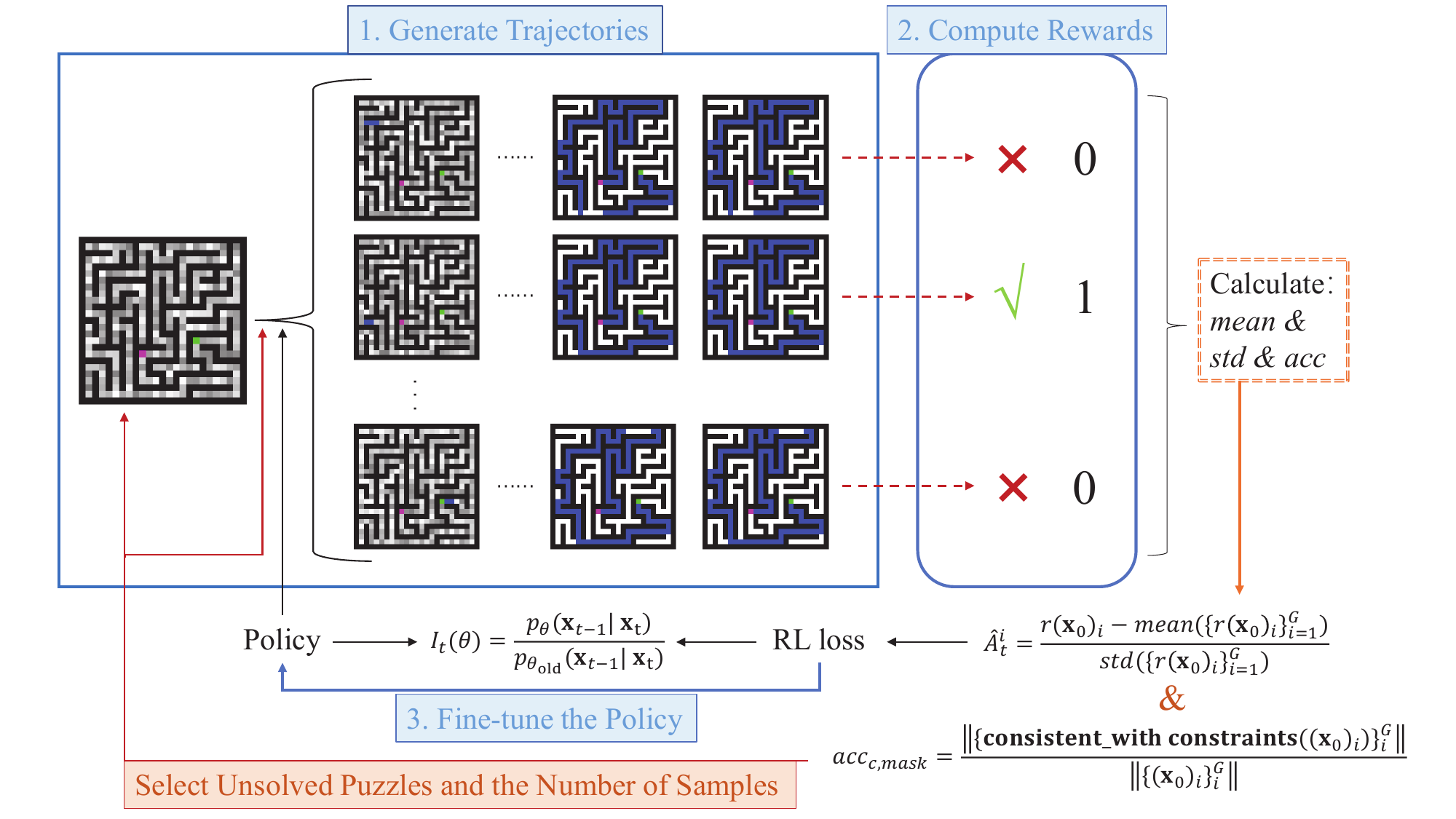}
  \caption{Overall workflow of DDReasoner's RL training.}
  \label{rl_pic}
\end{figure*}

\subsection{Markov Decision Processes and Reinforcement Learning}
Reinforcement learning (RL) is a branch of machine learning where an agent learns to make sequential decisions by interacting with an environment to maximize cumulative rewards \cite{Sutton1998}. It is typically modeled as a Markov Decision Process (MDP) defined by a tuple: $\mathcal{M} = (\mathcal{S}, \mathcal{A}, \rho_0, P, R)$ with states $ s \epsilon \mathcal{S}$, actions $a \epsilon \mathcal{A}$, initial state distribution $\rho_0$, transition probabilities $P$ and reward function $R$. At each timestep $t$,  the agent observes a state $ s_t \epsilon \mathcal{S}$, takes an action $a_t\sim\pi(a_t | s_t) \epsilon \mathcal{A}$, transitions to a new state $s_{t+1}\sim P(s_{t+1} | s_t,a_t)$ and receives a reward $R(s_t,a_t)$. As the agent acts in the MDP according to the policy $\pi$, it generates $\tau=(s_0, a_0,s_1, a_1, ...,s_T, a_T)$, a sequence of states and actions, i.e., trajectory. We aim to optimize the expected cumulative reward, discounted by a function $\gamma(\cdot)$, over trajectories sampled from its policy:
\begin{align}
    \mathcal{J}_{RL}=\mathbb{E}_{\tau\sim p(\tau|\pi)}\left[ \Sigma_{t=0}^T\gamma(t) R(s_t,a_t)\right] \label{mdp-function}
\end{align}

Among various approaches to RL, policy gradient methods directly optimize the policy $\pi$ by computing gradients of expected reward with respect to policy parameters. While standard approaches often suffered from high variance and instability, \citet{schulman2017proximalpolicyoptimizationalgorithms} proposed Proximal Policy Optimization (PPO) to improve training stability. PPO-CLIP is the most commonly used method, introducing a clipped surrogate objective to limit the deviation between the new and old policies during updates. The clipped objective is expressed as:
\begin{align}
    L^{\text{PPO}}(\theta) &= \mathbb{E}_t\left[ \min\left(I_t(\theta)
 \hat{A}_t,\text{clip}\left(I_t(\theta)
, 1 - \epsilon, 1 + \epsilon\right) \hat{A}_t \right) \right], \nonumber \\
& \text{where} \quad I_t(\theta) =\frac{\pi_\theta(a_t \mid s_t)}{\pi_{\theta_{\text{old}}}(a_t \mid s_t)}
\end{align}

Above, $\pi_\theta$ and $\pi_{\text{old}}$ is the current and previous policy respectively, and $\hat{A}_t$ is the advantage estimate which describes how much better or worse action $a_t$ is than what the current policy would average do in state $s_t$. More recently, Group Relative Policy Optimization (GRPO) \cite{deepseek-math} uses group statistics to derive advantages. For each question $q$, GRPO samples a group of outputs from the old policy and then computes the normalized reward in each group as the advantage to update the policy model.

\begin{table*}
  \centering
  \begin{tabular}{lllll}
    \toprule
    dataset name     & \# hints[\textit{min-max}]     & size  & \# of solutions  & main challenge \\
    \midrule
    big\ kaggle & 33.82 [29-37]  &100,000 & unique & scaling     \\
    minimal\ 17     & 17.00 [17-17] &  49,158 & unique &  minimal number of hints \\
   multiple\ sol     & 34.75 [34-35]   & 10,000    & multiple & 2 or more solutions  \\
   satnet\ data     & 36.22 [31-42]   & 10,000    & unique & -  \\
    \bottomrule
  \end{tabular}
  \caption{Summarization of Sudoku datasets. \textit{min, max} separately represent the minimum and maximum of filled cells in a board.}
  \label{dataset-table}
\end{table*}

\section{Methodology}

In this section, we introduce our diffusion-based pipeline for symbolic reasoning and our constraints-guided reinforcement learning technique to fine-tune the diffusion reasoner.

\subsection{Denoising Diffusion for Symbolic Reasoning}

We employ a diffusion model to perform symbolic reasoning, i.e., \textbf{DDReasoner}. After logical puzzles are represented as distributions in the continuous space (i.e., $p_\text{puzzle}(\textbf{x}_T)$), DDReasoner converts them to a probability distribution over possible complete solutions. It generates samples from a masked distribution $p(\textbf{x}_T) \triangleq mask \odot p_\text{puzzle}(\textbf{x}_T) + (1-mask)\odot\mathcal{N}(\textbf{x}_T;\textbf{0}, \textbf{I})$, where some positions (i.e., where $mask == 1$) are initially given as hints of the puzzle, and others are sampled from a Gaussian distribution. During the sampling process, observed hints remain unchanged, and other positions are denoised step-by-step to recover the distribution of a complete solution: 
\begin{align}
    p_\theta(\textbf{x}_{t-1}|\textbf{x}_t) = &\ \mathcal{N}(\textbf{x}_{t-1}|((1-mask) \odot \mu_t(\textbf{x}_t, \epsilon_\theta(\textbf{x}_t, t)) \nonumber \\
    & + mask \odot c), ((1-mask) \odot \sigma_t^2)\textbf{I}),
\end{align}
 in which $c\sim p_\text{puzzle}(\textbf{x}_T)$ is the known information from the puzzle. In the final step, $x_0$ in the continuous space is discretized to produce the predicted solution.
 
 We first use masked DDPM training to enable DDReasoner to preliminarily possess the capability to solve logical puzzles. We progressively add noise to unmasked positions in the complete solution distribution to obtain a noisy representation. DDReasoner is trained to predict the added noise:
\begin{align}
    \mathcal{L}_{\text{SL}}(\theta) =& \ \mathbb{E}_{\mathbf{x}_0, t, \mathbf{x}_t, c, mask} [\ \left\| \left( \boldsymbol{\tilde{\epsilon}}(\textbf{x}_0,t) - \right. \right. \nonumber \\ 
    &\left. \left. \boldsymbol{\epsilon}_\theta((mask \odot c +  (1-mask) \odot \mathbf{x}_t), t) \right) \right\|^2 ], 
\end{align}
where $c\sim p_\text{puzzle}(\textbf{x}_T)$ and $mask$ indicates the observed position in $c$. At inference time, we adopt a deterministic greedy sampling strategy: $\textbf{x}_{t-1} = mask \odot c +  (1-mask) \odot \mu_t(\textbf{x}_t, \epsilon_\theta(\textbf{x}_t, t)))$. The process of supervised learning and inference of DDReasoner is illustrated in Figure \ref{diffusion_pic}.

\subsection{Constraints-Guided Policy Optimization}
Following the supervised learning phase, we employ a flexible constraints-guided policy optimization technique for enabling the diffusion reasoner to internalize hard constraints.

\subsubsection{A Denoising Diffusion MDP}
\label{mdp}
Given a puzzle denoted by $\{c, mask\}$, the pretrained DDReasoner generates a sample distribution $p_\theta(\mathbf{x}_0|c,mask)$ through a fixed sampling process. Considering the sequential property of the denoising process, we regard it as a multi-step MDP following \citet{black2024trainingdiffusionmodelsreinforcement}, which is defined by:
\[s_t \triangleq (\textbf{x}_t, t) \ \ \ \ \  a_t\triangleq \textbf{x}_{t-1}  \ \ \ \ \     \pi(a_t | s_t)\triangleq p_\theta(\textbf{x}_{t-1} | \textbf{x}_t) \]
\[ P(s_{t+1}|s_t, a_t)\triangleq (\delta_{t-1},\delta_{\textbf{x}_{t-1}}) \]
$$ 
R(s_t,a_t)\triangleq \left\{
\begin{array}{ll}
r(\textbf{x}_0)& \ \ \    \text{if}\ \  t=0 \\
0 &  \ \ \   \text{otherwise}
\end{array}
\right.
$$
Above, $\delta_y$ is the Dirac delta distribution with non zero density only at $y$, and $r$ is the reward signal.

The denoising sequence consists of $\mathcal{T}$ timesteps, after which the process transitions to a termination state representating DDReasoner’s predicted solution to the given puzzle. The cumulative reward of each trajectory in this MDP is equal to $r(x_0)$, which aligns with the goal of DDReasoner's RL training. Therefore, in conjunction with Equation (\ref{mdp-function}), the objective function is defined as:
\begin{align}
    \mathcal{J}_\text{RL}=\mathbb{E}_{\textbf{x}_0 \sim p_\theta(\textbf{x}_0|c,mask)}[r(\textbf{x}_0)]
\end{align}

We use the sampling trajectories $\tau = (\textbf{x}_T, \textbf{x}_{T-1},..., \textbf{x}_0)$ collected during the sampling process to update the model parameters via gradient descent. To allow for multiple optimization steps per data collection round, we use an importance sampling estimator \cite{Kakade2002ApproximatelyOA} to estimate the policy gradient:
\begin{align}
    \nabla_\theta\mathcal{J}_{\text{RL}} = \mathbb{E} \left[ \sum_{t=0}^T \frac{p_\theta(\textbf{x}_{t-1} | \textbf{x}_t)}{p_{\theta_\text{old}}(\textbf{x}_{t-1} | \textbf{x}_t)} \nabla log p_\theta(\textbf{x}_{t-1} | \textbf{x}_t)r(\textbf{x}_0) \right]
\end{align}

\subsubsection{Constraint-Based Reward Modeling}
The consistency of neural predictions is crucial for neuro-symbolic learning tasks. To guide the generated solution to better satisfy hard logical constraints, we use the validity of the final generated solution as the outcome reward with the following rule:
\begin{align}
    r(\textbf{x}_0)=\left\{ 
    \begin{array}{ll}
    1 &    \text{if}\ \ \textbf{consistent\_with\_constraints}(\textbf{x}_0) \\
    0 &    \text{otherwise}
    \end{array}
    \right. ,
\end{align}
\begin{align}
\text{w}&\text{here }
    \textbf{consistent\_with\_constraints}(\textbf{x}_0)= \nonumber \\
    &\left\{ 
    \begin{array}{ll}
    \textbf{equivalent}(\textbf{Discretize}(\textbf{x}_0),a) &    \text{if}\ \ \textbf{unique}(a) \\
    \textbf{check\_rules}(\textbf{Discretize}(\textbf{x}_0)) &    \text{otherwise}
    \end{array}
    \right.
\end{align}

The ground-truth answer $a$ is acquired from a task-specific symbolic solver or computational algorithm in advance, so the computational cost of checking $\textbf{x}_0$ and $a$ is negligible if $a$ is unique. For tasks that may admit multiple consistent solutions, the computational complexity of this verification based on constraint rules is nevertheless substantially lower than that of solving the problem from scratch.

\subsubsection{Group-Based Dynamic Sampling}
Following \citet{black2024trainingdiffusionmodelsreinforcement} and \citet{deepseek-math}, we develop a group-based dynamic sampling method to optimize the diffusion reasoner's denoising process. For each puzzle representation $\{c, mask\}\ \epsilon\ D$, DDReasoner samples a group of trajectories $\{\tau_1,\tau_2,...,\tau_{G}\}$ from the old policy $p_{\theta_{\text{old}}}$. The cumulative reward of trajectory $\tau_i$ is equivalent to $r(\textbf{x}_0)_i$, so the estimated advantage of the $i\text{-th}$ trajectory is calculated by normalizing the group-level rewards $\{r(\textbf{x}_0)_i\}_{i=1}^G$:
\begin{align}
    \hat{A_t^i}={\frac{r(\textbf{x}_0)_i-mean(\{r(\textbf{x}_0)_i\}_{i=1}^G)}{std(\{r(\textbf{x}_0)_i\}_{i=1}^G)}},
\end{align}
which quantifies the relative level of constraint satisfaction of the $i\text{-th}$ trajectory compared to the average. During training, we record its average solved rate by calculating $acc_{c, mask} = \frac{\|\{\textbf{consistent\_with\_constraints}((\textbf{x}_0)_i)\}_i^G\|}{\|\{(\textbf{x}_0)_i\}_i^G\|}$. In the initial epoch, $G$ of each puzzle is set to $G_{\text{initial}}$; In the subsequent training iterations, DDReasoner dynamically adjusts both the selection of puzzles and the number of samples per puzzle based on the prior performance $\{acc_{c, mask}\}_D$. Specifically, we retain only the unsolved problems from the previous epoch. Additionally, at the start of a new epoch, we utilize the records from the previous round of training and adjust the number of samples allocated to each puzzle accordingly. For more challenging puzzles (i.e., those with lower solution accuracy $acc_{c, mask}$), we increase the number of samples (i.e., assign $G\ \epsilon [G_{\text{initial}}, \gamma*G_{\text{initial}}], \text{where }\gamma >1\text{ is a scaling factor}$) to encourage the model to explore a broader range of potential solution paths, with the goal of discovering trajectories that successfully solve the puzzle.

To prevent model overfitting and the occurrence of reward hacking, we introduce a clipping operation to ensure that $p_\theta$ does not deviate too far from $p_{\theta_\text{old}}$, forcing the policy update ratio $I_t(\theta)=\frac{p_\theta(\textbf{x}_{t-1} | \textbf{x}_t)}{p_{\theta_\text{old}}(\textbf{x}_{t-1} | \textbf{x}_t)}$ is constrained within a threshold range, which is commonly used in PPO. 

Finally, the clipped objective is expressed as follows:
\begin{align}
\label{obj}
    & \mathcal{J}_\text{RL}(\theta)=\\
    & \mathbb{E}_{t} \left[\sum_{i=1}^{G} \min\left(I_t^i(\theta)
 \hat{A}_t^i, \text{clip}\left(I_t^i(\theta)
, 1 - \epsilon, 1 + \epsilon\right) \hat{A}_t^i \right) \right] \nonumber
\end{align}
Above, $I_t^i(\theta)$ and $\hat{A}_t^i$ have been defined, and $G$ is decided based on the specific puzzle.

The overall pipeline of the policy optimization algorithm is summarized in Figure \ref{rl_pic} and Algorithm 1 in Appendix B.2.

\begin{table*}
  \centering
  \begin{tabular}{lllll}
    \toprule
    model  & big\ kaggle  & minimal\ 17  & multiple\ sol & satnet\ data \\
    \midrule
    Baseline* (\citet{cornelio_2023_NASR}) & 62.68 & 0.00 & 46.70 &  24.00 \\
    RNN+PseudoSL (\citet{ahmed2023a}) & —— & —— & —— & 28.20 \\
    DDReasoner-SL & 78.19 & 8.03  &  96.30 & 70.60 \\
    \midrule
    DDReasoner-RL (Ours) & \textbf{97.79}$\uparrow$ & \textbf{18.25}$\uparrow$  &  \textbf{100.00}$\uparrow$ & \textbf{92.60}$\uparrow$ \\
    \bottomrule
  \end{tabular}
  \caption{Our experimental results on Sudoku. *Baseline is the sub-module \textbf{SolverNN} in NASR\cite{cornelio_2023_NASR}. It is a Transformer model with a linear layer mapping the input distribution to the probability distribution over the possible complete solutions. Its inputs and labels are consistent with our experiment, so we choose it as the Baseline. The metric used in the table for each dataset is the percentage of completely correct predicted boards, and no partial credit is given for getting individual digits correct. \textbf{——} in the table means that the code is inaccessible, so we were unable to reproduce its results on other datasets.}
  \label{sudoku_result}
\end{table*}

\begin{table*}
  \centering
  \begin{tabular}{llllll}
    \toprule
    \multicolumn{2}{c}{model} & \multicolumn{4}{c}{maze grid size} \\
    \cmidrule(r){1-2}
    \cmidrule(r){3-6}
    name & \# parameters & $5 \times 5$ & $10 \times 10$  & $15 \times 15$ & $20 \times 20$\\
    \midrule
    Baseline* (\citet{nolte2024transformersnavigatemazesmultistep}) & 8M & \textbf{100.00} & \textbf{100.00} & \textbf{100.00} & \textbf{100.00}\\
    DDReasoner-SL & 5.3M & 95.45 &  70.95 & 51.35  &  43.30 \\
    \midrule
    DDReasoner-RL (Ours) & 5.3M & \textbf{100.00}$\uparrow$ & \textbf{100.00}$\uparrow$ & \textbf{100.00}$\uparrow$  &  \textbf{100.00}$\uparrow$ \\
    \bottomrule
  \end{tabular}
  \caption{Our experimental results on Maze. *Baseline is tailored for Maze navigation based on Transformer and $\text{MLM-}\mathcal{U}$ \cite{kitouni2024the}. It uses the same dataset configuration as ours, and can perfectly solve mazes of grid sizes up to $20 \times 20$.}
  \label{maze_result}
\end{table*}

\section{Experiment}
\label{exp}
\subsection{Tasks}
\label{task}
While these constraint satisfaction problems are trivial for computers to solve based on logical rules and programming, they are all challenging for neural networks. In the following part, we provide a brief introduction to each task; more experimental details can be found in Appendix B.1.

\paragraph{Sudoku}
Sudoku is a classical logic-based puzzle game, in which a player must fill in a 9 × 9 partially-filled board such that each row, each column, and each of nine 3 × 3 subgrids contains exactly one of each number from 1 through 9. Following the setup of \citet{cornelio_2023_NASR}, we consider three public Sudoku datasets: 1) big kaggle, a subset of a bigger dataset containing 1 million boards hosted on Kaggle; 2) minimal 17, a dataset of minimal Sudoku boards, with only 17 clues (the minimum of hints in a Sudoku board leading to a unique solution); and 3) multiple sol, a dataset of Sudoku boards with two or more solutions. 4) Finally, we considered satnet\_data, released by \citet{DBLP:journals/corr/abs-1905-12149}. The information of these datasets is summarized in Table \ref{dataset-table}. For each Sudoku dataset, we use Prolog or a brute-force algorithm to obtain solutions (if unlabeled), and split it into a training set with 90\% examples and a test set with 10\% examples.

\paragraph{Maze}
Maze is a structured navigation task. The player begins at a starting point and must navigate a connected path to reach a specified endpoint. Movement is generally allowed in four directions: up, down, left, and right, and walls or obstacles cannot be crossed. Our Maze datasets are generated using the configuration of \citet{maze-dataset}. For Maze grid sizes of $5 \times 5$, $10 \times 10$, $15 \times 15$ and $20 \times 20$, we separately generate 20'000 puzzles and their corresponding solution pathways, with 18'000 pairs as training set and 2'000 pairs as test set. Each puzzle is generated by Depth-First Search (DFS) and is acyclic, so its solution is unique.

\paragraph{Simple Path Prediction \& Preference Learning}
We also adopt the neuro-symbolic benchmarks of \citet{pmlr-v80-xu18h}. 1) Simple Path Prediction: Given a source and destination node in a 4-by-4 unweighted grid $G = (V, E)$, we need to find the shortest unweighted path connecting them. Formally, the known condition is a binary vector of length $|V|+|E|$, where $|V|$ indicates the source and destination nodes and $|E|$ indicates which edges are removed, thus obtaining a subgraph $G' \subseteq G$. The corresponding solution is a binary vector of length $|E|$ indicating which edges are in the shortest path. The dataset consists of 1610 such examples. 2) Preference Learning: Given a user’s ranking over a subset of items, we want to predict the user’s ranking over the remaining items. We use preference ranking data over 10 types of sushi, taking the ordering over 6 types of sushi as the known condition to predict the ordering  (a strict total order) over the remaining 4 types. We use preference ranking data over 10 types of sushi for 4926 individuals. For both tasks, we split the data 60\%/20\%/20\% into train/validation/test split.

\paragraph{Minimum-Cost Path Finding}
Given a weighted $k \times k$ grid, we need to predict the path with minimum cost from the upper left to the lower right vertices. The complete solution can be represented as a $k \times k$ matrix. Notably, the minimum-cost path is possibly not unique, i.e., for a puzzle, two or more minimum-cost paths may exist, and they are all considered correct. Adopting datasets from \citet{Pogančić2020Differentiation}, we consider $k=12,18,24,30$, and use 10'000 examples as the training set and 1'000 examples as the test set.

\subsection{Experiment Setup}
We allocate a certain percentage of training data for the SL phase. In the RL phase, we continue training from the SL checkpoint using the whole training set. Detailed configurations of both training phases are provided in Appendix B.3. 

\subsection{Main Results}

DDReasoner-SL is developed on a basic diffusion model and partial data via supervised learning, a stage where other efforts have been directed \cite{Du_2024_ICML, zhang2025tscendtesttimescalablemctsenhanced, 10.1145/3637528.3671783}. We do not focus on elaborate supervised learning nor rely on test-time scaling. Instead, we pay more attention on employing RL techniques to directly internalize logical rules and specific constraints into DDReasoner's capabilities while training, thus allowing for better single-pass performance at inference time. In following results, solutions produced by DDReasoner are obtained via an efficient single-pass deterministic sampling.

\paragraph{Sudoku}
As demonstrated in Table \ref{sudoku_result}, we outperform the purely neural network Baseline on all four datasets. For the most challenging dataset (minimal\_17 with only 17 hints each Sudoku board), the baseline SolverNN cannot produce any completely correct solution board, while DDReasoner is capable of yielding a number of consistent predictions, and further strengthens the reasoning ability through RL training. Particularly on the dataset featuring multiple solutions that satisfy the hard constraints (multiple\_sol), our RL method based on constraint satisfaction verification empowers DDReasoner to comprehensively grasp this rule, thereby reaching perfect accuracy. These results show that once DDReasoner is equipped with a preliminary capability for generating consistent solutions, we can further refine this particular capability via our policy optimization algorithm.

\begin{table}
  \centering
  \begin{tabular}{llll}
    \toprule
    model  & exact  & hamming & consistent \\
    \midrule
    \small{MLP} & 5.62 & 85.91 & 6.99 \\
    \small{I. MLP + Semantic Loss} & 28.51 & 83.14 & 69.89 \\
    \small{II. MLP + NESYENT} & 30.10 & 83.00 & 91.60 \\
    \small{III. MLP + SPL} & 37.60 & 88.50 & \textbf{100.00} \\
    \midrule
    \small{DDReasoner-SL} & 66.46 & 92.71  &  74.53 \\
    \small{DDReasoner-RL (Ours)} & \textbf{70.81}$\uparrow$ & \textbf{93.96}$\uparrow$ & 80.12$\uparrow$ \\
    \bottomrule
  \end{tabular}
  \caption{Our experimental results on simple path prediction. I, II and III are taken from \citet{pmlr-v80-xu18h}, \citet{Ahmed2022NeuroSymbolicER} and \citet{NEURIPS2022_c182ec59}, respectively. I and II used constraint-based losses, and III added a constraint layer.}
  \label{grid_result}
\end{table}

\begin{table}
  \centering
  \begin{tabular}{llll}
    \toprule
    model  & exact  & hamming & consistent \\
    \midrule
    \small{MLP} & 1.01 & 75.78 & 2.72 \\
    \small{I. MLP + Semantic Loss} & 13.59 & 72.43 & 55.28 \\
    \small{II. MLP + NESYENT}  & 18.20 & 71.50 & 96.00 \\
    \small{III. MLP + SPL} & 20.80 & 72.40 & \textbf{100.00} \\
    \midrule
    \small{DDReasoner-SL} & 21.02 & 78.15 &  99.80 \\
    \small{DDReasoner-RL (Ours)} & \textbf{22.54}$\uparrow$ & \textbf{78.57}$\uparrow$ & \textbf{100.00}$\uparrow$\\
    \bottomrule
  \end{tabular}
  \caption{Our experimental results on preference learning. }
  \label{preference_result}
\end{table}

\paragraph{Maze}

As demonstrated in Table \ref{maze_result}, we achieve 100\% accuracy for mazes of grid sizes up to $20 \times 20$ via our policy optimization algorithm. Figure 6 in Appendix B.4 shows a performance comparison between our DDReasoner-RL and Baseline on $5 \times 5$ mazes. We observe that this perfect performance of our DDReasoner-RL is accompanied by approximately 1.51x more parameters-efficient, 1.39x more data-efficient, and 2.33x more time-efficient than Baseline, respectively. It underscores that training DDReasoner with our RL method can effectively elicit its potential reasoning capabilities while ensuring notable efficiency.

\paragraph{Simple Path Prediction \& Preference Learning}
As demonstrated in Table \ref{grid_result} and Table \ref{preference_result}, our DDReasoner-RL surpasses previous loss-based methods on the two tasks in terms of the accuracy of exact match and hamming distance, while also achieving a high level of consistency satisfaction.

\paragraph{Minimum-Cost Path Finding}
As demonstrated in Table \ref{warcraft_result}, for cases where $k=12,18,24,30$, the use of RL yielded notable enhancements to both exact and consistent. 
\begin{table}
  \centering
  \begin{tabular}{llllll}
  \toprule
    model  & &  k=12& k=18 &k=24 & k=30 \\
    \midrule
    \multirow{2}{*}{DDReasoner-SL} & I & 88.70 & 81.90 & 75.00 & 70.60 \\
    & II & 98.70 & 97.70 & 96.60 & 93.20 \\
    \midrule
    \multirow{2}{2.5cm}{DDReasoner-RL\\(Ours)} & I & \textbf{90.40} & \textbf{86.20} & \textbf{81.30} & \textbf{80.10} \\
    & II & \textbf{99.10} & \textbf{99.10} & \textbf{99.10} & \textbf{98.40} \\
    \hline
\end{tabular}
\caption{Our experimental results on minimum-cost path finding. I and II denote the percentage of finding the shortest path and a connected path, respectively.}
\label{warcraft_result}
\end{table}

\subsection{Ablation Studies and Overall Analysis}
\begin{table}
  \centering
  \begin{tabular}{ccccc}
    \toprule
    task & \multicolumn{2}{c}{\small{DDReasoner-SL*}} & \multicolumn{2}{c}{\small{DDReasoner-RL}} \\
    \cmidrule(r){2-3}
    \cmidrule(r){4-5}
    & train & test & train & test \\
    \midrule
    \small{Sudoku (big\ kaggle)} & 99.30 & 82.82 & 99.97 & \textbf{97.79} \\
    \small{Sudoku (minimal\ 17)} & 99.88 & 9.48 & 99.04 & \textbf{18.25} \\
    \small{Sudoku (multiple\ sol)} & 95.81 & 97.20 & 100.00 & \textbf{100.00}\\
    \small{Sudoku (satnet\ data)} & 99.87 & 71.70 & 99.87 & \textbf{92.60} \\
    \small{Maze ($5 \times 5$)} & 99.90 & 99.85 & 100.00 & \textbf{100.00} \\
    \small{Maze ($10 \times 10$)} &  95.08 & 91.35  & 99.98 & \textbf{100.00} \\
    \small{Maze ($15 \times 15$)} & 59.94 &  58.10 & 100.00 & \textbf{100.00} \\
    \small{Maze ($20 \times 20$)} & 62.77  & 58.35 & 99.99 & \textbf{100.00} \\
    \bottomrule
  \end{tabular}
  \caption{Partial results of ablation studies. DDReasoner-SL* in this table is obtained by scaling up supervised learning.}
  \label{ablation}
\end{table}
To justify the adoption of our RL method, we take the SL checkpoint that is prepared for subsequent RL training, and scale up supervised learning using unsolved puzzles of the entire training set. As shown in Table \ref{ablation} and Table 12,13,14 in Appendix B.4, utilizing the constraints-based reward signal to guide DDReasoner's policy optimization has refined the SL checkpoint's ability to adhere to logical constraints to a large extent. Furthermore, additional experiments on the generalization ability of our DDReasoner-RL across different Sudoku datasets and varying Maze grid sizes are provided in Table 10 and 11 in Appendix B.4.

At inference time, DDReasoner-RL can generate predicted solutions in batch for puzzles in the test set within a few seconds using 4 CPUs and 1 NVIDIA H100 GPU.

\section{Conclusion and Future Work}
We use a diffusion model as a symbolic logical reasoner (i.e., DDReasoner). We establish DDReasoner's preliminary reasoning skills via supervised learning, and then, use reinforcement learning to finetune DDReasoner via formulating the denoising process as a Markov decision process. Our reward signal adopts the constraint satisfaction verification to prioritize logically consistent outputs. Our experimental results demonstrate that our integration of RL ultimately enables the diffusion model to be a proficient reasoner capable of addressing diverse symbolic reasoning tasks based on constraint satisfaction. While our current tasks focus on the low-dimensional state space, future work could involve applying this method to high-dimensional problems, e.g., visual Sudoku and visual generation related to physical consistency. The sole modification is the incorporation of a separate neural network to handle the discretization of the final denoising state and the subsequent reward calculation.

\bibliography{aaai2026}


\newpage
\section{Appendix}
\section{A \ \ \ \ Code}
Our source code is provided in

https://github.com/dd88s87/DDReasoner.

\section{B \ \ \ \ Additional Details for Experiments}
\subsection{1 \ \ Detailed Implementation of Each Task}
DDReasoner generates samples from a masked distribution:
\begin{equation}
    p(\textbf{x}_T) \triangleq mask \odot p_\text{puzzle}(\textbf{x}_T)\ +\ (1- mask)\odot\mathcal{N}(\textbf{x}_T;\textbf{0}, \textbf{I}), \nonumber
\end{equation}

Above, $p_\text{puzzle}(\textbf{x}_T)$ is a distribution in a continuous space that represents a puzzle in the form of one-hot representations, and $mask$ is a binary mask where $mask==1$ indicates partial observations in the puzzle. These settings vary across different tasks and even between individual puzzles, and we will provide an introduction in the following section.

\subsubsection{Sudoku}
A Sudoku board is valid if the entries of each row, column, and $3 \times 3$ square are unique, i.e., each number from 1 to 9 appears exactly once. Some examples of both valid and invalid Sudoku solutions are shown in Figure \ref{examples}.

To convert a Sudoku board into the input of DDReasoner, each cell in the $9 \times 9$  grid is represented as a 9-dimensional one-hot vector, indicating possible digits from 1 to 9, as shown in Figure \ref{onehot}(a). A partially filled Sudoku puzzle is thus treated as a representation in the continuous space. Known values are fixed as one-hot vectors, and unknown positions are transformed into distributions over possible digits. Therefore, the final denoising state is a full $9 \times 9$ board where each cell contains a 9-dimensional probability distribution over digits 1 to 9. Finally, the discretization operation consists of applying an \textit{argmax} function over the 9 dimensions for each position, which converts the output back into its corresponding digit form.

\subsubsection{Maze}
The Maze game is a structured navigational task. The player begins at a designated starting point and must navigate a connected path to reach a specified endpoint. Movement is restricted to valid passages, and walls or obstacles cannot be crossed. Players are generally allowed to move in four directions: up, down, left, and right. The maze board consists of a series of interconnected pathways, designed with intentional complexity to challenge spatial reasoning and decision-making skills.

To convert a Maze grid into the input for DDReasoner, the $N \times N$ grid is represented as a one-hot format with the size of $2 \times N \times N$. Each cell $(i,j)$ is denoted by a 2-dimensional hot vector, where the $1\text{-st}$ dimension indicates the "wall" and the $2\text{-rd}$ indicates the "passage". We encode the start point and end point by our mask. That is, we directly set the start and end point locations to be on the solution path by fixing their corresponding values in the $2\text{-rd}$ dimension to 1. These specific locations are then masked to ensure they serve as immutable conditions guiding solution generation during the denoising process. In addition, all positions representing walls within the maze are also masked to remain constant, functioning as pre-observed conditional information. An illustrative example is provided in Figure \ref{onehot}(b). Discretization of the final denoising state is performed by binarizing the predictive output for each location in the $2\text{-rd}$ dimension. Specifically, a position is classified as belonging to the predicted path if its value exceeds a threshold of 0.5.

\subsubsection{Simple Path Prediction \& Preference Learning}
\label{simple and preference}
\paragraph{Simple Path Prediction}
We provide DDReasoner with the input sequence of length $|V|+|E|$. In the initial representation, start and end points are set to 1, while all remaining vertices and edges are set to 0. Furthermore, a mask (with values set to True) is applied to all vertex positions and to positions corresponding to deleted edges, thereby designating these as known, fixed information. The final discretization is performed by binarizing the predictive output.

\paragraph{Preference Learning}
We represent a user's preferences for 10 sushi items as a $10 \times 1 \times 10$ sequence. In this representation, the 10 channels for each item constitute a one-hot encoding of its rank among all 10 items. For 6 sushi items with pre-determined ranks, their one-hot representations are masked to serve as fixed conditions. DDReasoner's task is then to predict the one-hot encoded ranks for the remaining 4 items. The final discretization is performed by applying an \textit{argmax} function over the predictive output.

\subsubsection{Minimum-Cost Path Finding}
For a $k \times k$ grid in this task, it is encoded into a $2 \times k \times k$ representation. The first channel, representing the complete map's cost, is entirely masked as it constitutes pre-known conditional information. The second channel is designated for the predicted minimum-cost path; it is initialized by assigning a value of 1 to the start and end vertices (the map's top-leftmost and bottom-rightmost) and 0 to all other locations. After inference, nodes on the resultant predicted path are updated to 1. Crucially, in this second channel, the start and end points are themselves masked to ensure they are treated as fixed anchors for the path under prediction, with DDReasoner subsequently determining the intermediate nodes progressively. An illustrative example is provided in Figure \ref{onehot}(c). The final discretization is performed by binarizing the predictive output in the $2\text{-rd}$ dimension.

\setcounter{figure}{2}
\setcounter{table}{7}
\begin{figure}
  \centering
  \includegraphics[width=1.0\linewidth]{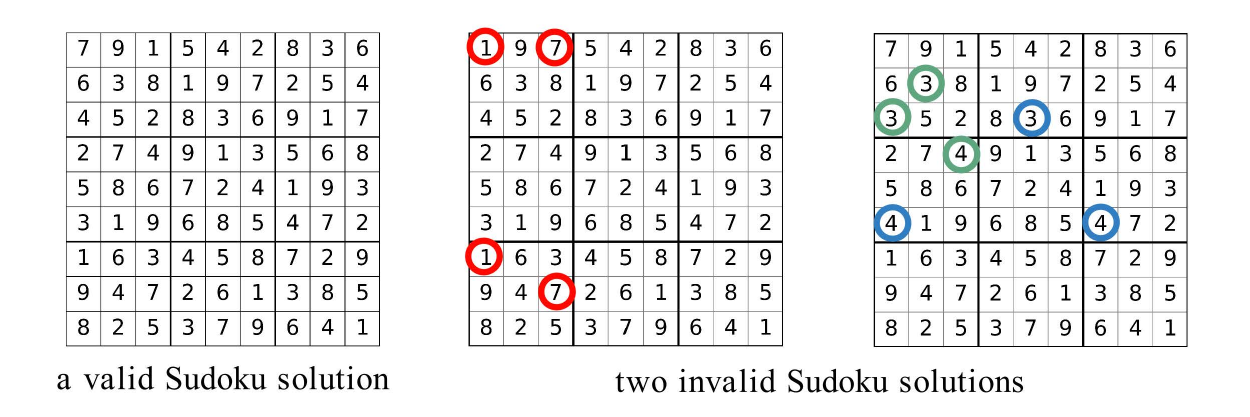}
  \caption{Examples of a valid Sudoku solution and two invalid Sudoku solutions. Red, blue, and green circles mark conflicts of column, row, and subgrid, respectively.}
  \label{examples}
\end{figure}

\begin{figure*}
  \centering
  \includegraphics[width=0.65\linewidth]{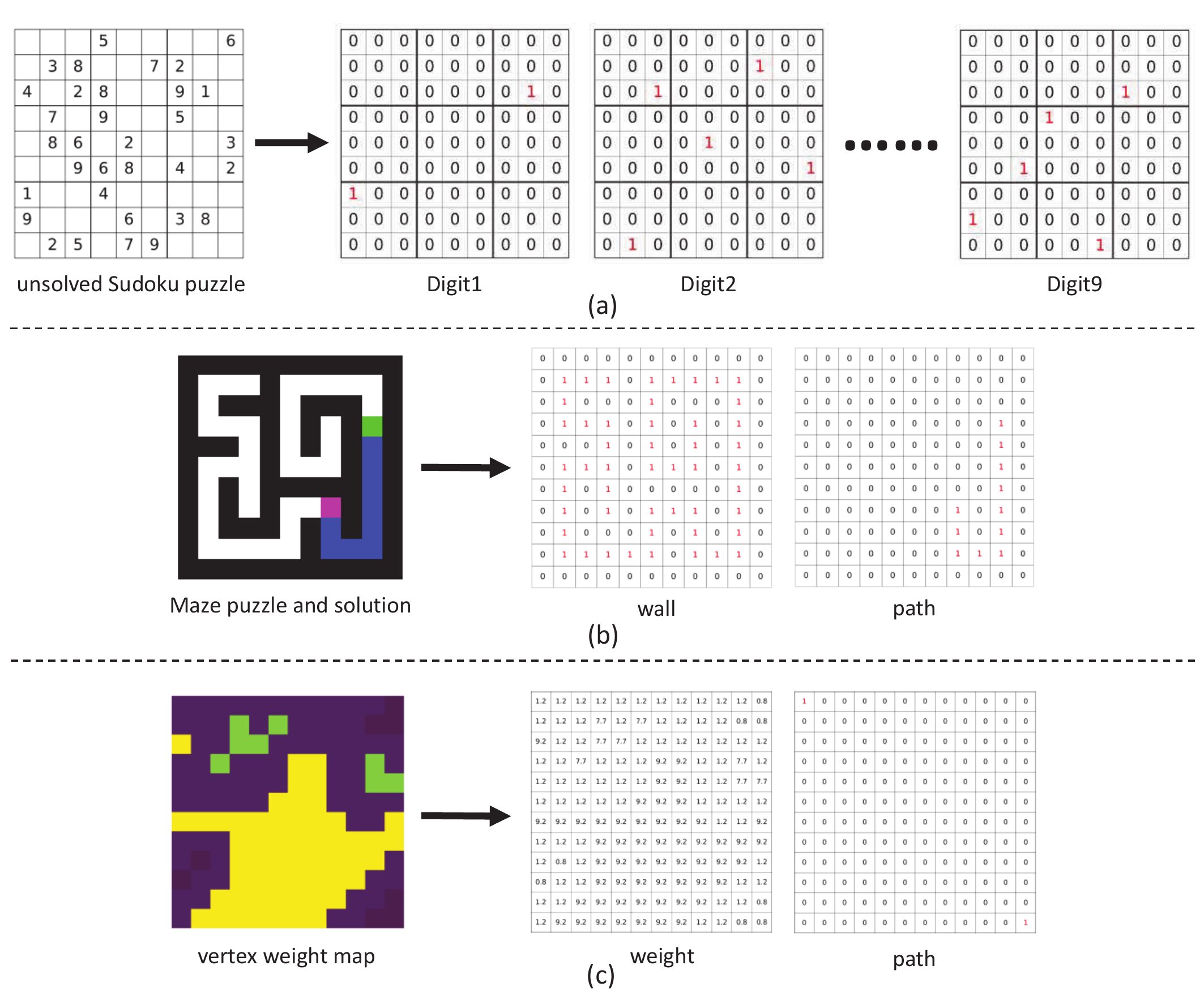}
  \caption{(a) An example of one-hot representation of Sudoku puzzles; (b) An example of one-hot representation of $5 \times 5$ Maze grids; (c) An example of one-hot representation of vertex weight grids.}
  \label{onehot}
\end{figure*}
\subsection{2 \ \ Detailed implementation of DDReasoner}
For each task, we set the number of timesteps in the diffusion model is 20. 

The noise prediction model each step is implemented as a modified U-Net-based architecture, where all down-sampling and up-sampling operations are removed. Instead, the network maintains a constant spatial resolution throughout, which is particularly important for preserving positional consistency in structured prediction tasks in our work.

During the RL training phase for DDReasoner, we use the policy optimization algorithm in Algorithm \ref{alg}. The detailed description of its entire workflow is as follows:

The workflow for each epoch can be divided into four parts: 
\begin{enumerate}
    \item \textbf{Select data}: From the second epoch onwards, the remaining challenging data is then processed in mini-batches, in which puzzles are ranked in an ascending order according to their $acc_{c,mask}$ in the previous epoch. Problems that rank lower by this metric are then sampled more intensively during the current epoch.
    \item \textbf{Sample trajectories}: The pre-trained model $p_{\theta_{SL}}$ is used to generate $G$ solution trajectories, capturing the hidden layers of intermediate steps $\textbf{x}_t$ across $T$ steps as well as the final predicted solution $\textbf{x}_0$.
    \item \textbf{Calculate rewards}: The reward for each generated solution $\textbf{x}_0$ is primarily ascertained by its adherence to the problem's logical hard constraints. 
    \item \textbf{Fine-tune with RL methods}: The rewards collected in each trajectory $i$ are used to calculate the Advantage $\hat{A}_t^i$, while the importance sampling ratio $I^i_t$ is calculated using the hidden layers $\textbf{x}_t^i$ at each step $t$. The fine-tuned model $p_{\theta_{old}}$ is then updated.
\end{enumerate}
At the end of each epoch, the updated model $p_\theta$ is used to perform new sampling, facilitating an online training process.

\begin{algorithm*} 
	\caption{Constraints-Guided Policy Optimization for Diffusion Reasoner} 
	\label{alg} 
	\begin{algorithmic}
		\REQUIRE pretrained \textbf{DDReasoner} $p_{\theta_\text{SL}}$, dataset $\mathcal{D}=\{(c_i, mask_i,idx_i)\}_{i=1}^D$, the number of epochs $N$, the number of timesteps $T$, the reward function \textbf{consistent\_with\_constraints}, the initial number of samples per group $G_{\text{initial}}$, a scaling factor $\gamma\ (\gamma >1)$.
        \STATE Let $G_{j,k}$ denote the number of sampling trajectories of the puzzle with $idx_i==k$ at epoch $j$, and $Acc_{j,k}$ denote the average solved rate of the puzzle with $idx_i==k$ at epoch $j$.
        \STATE Initialize $p_\theta=p_{\theta_\text{old}}=p_{\theta_\text{SL}}, \mathcal{D}_1=\mathcal{D}$
        \FOR{$j=1$ to $N$}
            \STATE Iteratively collect a batch of logical puzzles $\{(c_i, mask_i,idx_i)\}_{i=1}^B \subseteq \mathcal{D}_j$.
            \IF{$j==1$}
            \STATE $G_{j,idx_i} \gets G_{\text{initial}}\ \ \forall\ i\ \epsilon\ [1, B]$
            \ELSE
            \STATE Rank $B$ puzzles in ascending order on $Acc_{j-1,idx_i}$ and assign $G_{j,idx_i}\ \epsilon\ [G_{\text{initial}}, \gamma*G_{\text{initial}}]$.
            \STATE \% For puzzles $i$ ranked earlier in the batch, a larger $G_{j,idx_i}$ is used.
            \ENDIF
            \STATE Collect $G_{j,idx_i}$ trajectories from $p_\theta$: $\small{ \ \forall\ i\ \epsilon\ [1, B]}$ \\
            \small{$\{(\textbf{x}_T, \textbf{x}_{T-1},..., \textbf{x}_0)\sim p_\theta(\textbf{x}_T)p_\theta(\textbf{x}_{T-1}|\textbf{x}_T)...p_\theta(\textbf{x}_{0}|\textbf{x}_1)\}$} \\ 
            \STATE Compute Constraints-based reward with the reward function $\textbf{consistent\_with\_constraints}$
            \STATE Compute the policy gradient for each timestep $t$ and each trajectory, update $\theta$: $p_{\theta_\text{old}} \gets p_\theta$ , and training dataset: \ \small{$ \mathcal{D}_{j+1} \gets \mathcal{D}_j\ -\ \{(c_i, mask_i,idx_i)|\ \forall\ i\ \epsilon\ [1, B] \land  Acc_{j,idx_i}=1\}$}
        \ENDFOR
	\ENSURE $\text{Fine-tuned \textbf{DDReasoner}}\ \ p_\theta$ 
	\end{algorithmic} 
\end{algorithm*}

For practical purposes, the detailed experimental configuration involves:
\begin{enumerate}
    \item In \textbf{step 1}, we set the hyperparameter $G_\text{initial}$ to 64 and the scaling factor $\gamma$ to 4. During the initial epoch, each puzzle is sampled 64 times. For subsequent epochs, the number of samples per puzzle is dynamically adjusted within the range of $ [64, 256]$. To prevent GPU memory overload and ensure training efficiency, we establish three discrete sampling tiers within this range: 64, 128, and 256. Based on each puzzle's rank within a batch, puzzles in the 0-1/8th rank interval are allocated 256 samples, those in the 1/8th-3/8th interval receive 128 samples, and the remaining puzzles are sampled 64 times.
    \item  In \textbf{step 3}, to enhance computational efficiency and training speed, rewards for all tasks are derived from direct comparison with the ground truth, except for Sudoku (multiple\_sol) and Minimum-Cost Path Finding. Given that puzzles in tasks Sudoku (multiple\_sol) and Minimum-Cost Path Finding can admit multiple valid solutions satisfying all constraints, a rule-based checking method is employed for reward calculation during their training phases.
\end{enumerate}

\subsection{3 \ \ Training Details}
\label{train_details}
The amount of data used for supervised learning varies depending on the difficulty level of the specific task.
\begin{table}
  \centering
  \begin{tabular}{llll}
    \toprule
    task     & batch size  &  \# samples \\
    \midrule
    Sudoku (big\_kaggle) & 1024   &  70'000    \\
    Sudoku (minimal\_17) & 1024   &  44'242    \\
    Sudoku (multiple\_sol) & 1024 &  7'000    \\
    Sudoku (satnet\_data) & 1024 &  9'000    \\
    Maze ($5 \times 5$, $10 \times 10$)     & 1024 & 18'000      \\
    Maze ($15 \times 15$)     & 512  & 18'000      \\
    Maze ($20 \times 20$)     & 256  & 18'000      \\
    simple path prediction     & 1024   & 966  \\
    preference learning     & 1024   & 2'955  \\
    minimum-cost path finding   & 512  & 10'000  \\
    \bottomrule
  \end{tabular}
  \caption{Training details of SL phase}
  \label{training-table}
\end{table}
During supervised learning, we used the AdamW optimizer and set the learning rate to $3 \times 10^{-5}$ to train the diffusion model for at most 5000 epochs. The hyperparameters of AdamW optimizer were set as follows: $\beta_1 = 0.9$, $\beta_2 = 0.999$, $\epsilon = 1 \times 10^{-8}$, and weight decay coefficient $= 1 \times 10^{-4}$. 
For different tasks, the detailed configuration is in Table \ref{training-table}. 

The duration of supervised learning for each task is approximately $1 - 7$ hours using 1 NVIDIA H100 GPU and 15 CPUs. The specific training time varies depending on the setup of the training batch size and the number of training samples.

After the supervised learning reaches a certain stage, it tends to suffer from overfitting, so we turn to the RL phase to further enhance DDReasoner's performance based on the SL checkpoint.

\begin{table*}
  \centering
  \begin{tabular}{lll}
    \toprule
    task     & sample per batch  &  learning rate \\
    \midrule
    Sudoku (big\_kaggle, multiple\_sol, satnet\_data) & 64  & $1 \times 10^{-6}$   \\
    Sudoku (minimal\_17) &  32  & $1 \times 10^{-6}$  \\
    Maze ($5 \times 5$)     & 64 & $5 \times 10^{-7}$  \\
    Maze ($10 \times 10$)    & 8 & $1 \times 10^{-6}$  \\
    Maze ($15 \times 15$)     & 4 & $3 \times 10^{-6}$  \\
    Maze ($20 \times 20$)     & 2 & $3 \times 10^{-6}$  \\
    Simple path prediction     & 32   & $1 \times 10^{-5}$  \\
    Preference learning     & 64   & $1 \times 10^{-6}$ \\
    minimum-cost path finding ($12 \times 12$)     & 64   & $1 \times 10^{-6}$  \\
    minimum-cost path finding ($18 \times 18$)     & 16   & $1 \times 10^{-6}$  \\
    minimum-cost path finding ($24 \times 24$)     & 8   & $1 \times 10^{-6}$  \\
    minimum-cost path finding ($30 \times 30$)     & 4   & $1 \times 10^{-6}$  \\
    \bottomrule
  \end{tabular}
  \caption{Training details of reinforcement learning phase}
  \label{rl-table}
\end{table*}

\begin{figure*}
  \centering
  \includegraphics[width=1.0\linewidth]{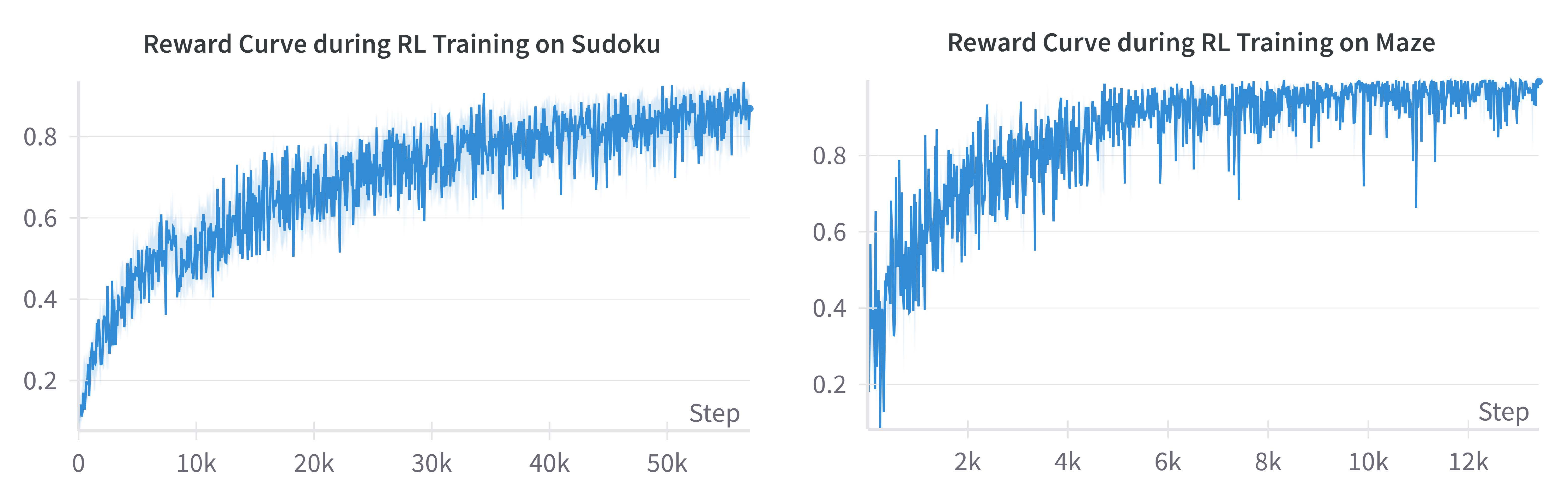}
  \caption{Examples of reward curve during RL training on Sudoku and Maze.}
  \label{reward_curve}
\end{figure*}

In the RL phase, we also adopted the AdamW optimizer with the same hyperparameters setting with the SL phase, and trained the diffusion model with the whole training set for at most 50 epochs. In this course, an early stopping criterion is employed: if the average reward on the whole training set (including data points that were fully correct and filtered from the training data) shows no improvement for 5 consecutive epochs or if all training data has been filtered out, early stopping is triggered to terminate the RL training. The detailed configuration is in Table \ref{rl-table} for different tasks.

The RL training time for each task is approximately $2 - 30$ hours using 1 NVIDIA H100 GPU and 15 CPUs. The specific training time varies depending on the setup of the number of samples per batch and the number of training samples.

Figure \ref{reward_curve} shows two examples of the reward curve during the RL training process.

\subsection{4 \ \ Addition Experiment Results}
\label{additional_results}
\subsubsection{Sudoku}
\paragraph{Experiments across Multiple Datasets}
To validate the impact of training with RL methods on the generalization ability of DDReasoner, we conducted the following cross-dataset experiments: DDReasoner was trained on Dataset \textbf{A} and tested on Dataset \textbf{B}.

As demonstrated in the Table \ref{cross_dataset}, after being fully trained on Dataset \textbf{A} using supervised learning, DDReasoner is highly prone to overfitting to that dataset, which in turn negatively affects their performance on Dataset \textbf{B}. However, training with our RL method leads to better generalization on Dataset \textbf{B}.

\begin{table*}
  \centering
    \begin{tabular}{llll}
    \toprule
    set  & DDReasoner-SL  & DDReasoner-SL* & DDReasoner-RL \\
    \midrule
    \textbf{A}: big\_kaggle, \textbf{B}: multiple\_sol & 3.30 & 3.40 & 23.60 \\
    \textbf{A}: big\_kaggle, \textbf{B}: minimal\_17 & 0.00 & 0.00 &  0.02 \\
    \textbf{A}: big\_kaggle, \textbf{B}: satnet\_data & 89.20 & 91.30 &  99.00 \\
    \textbf{A}: satnet\_data, \textbf{B}: big\_kaggle & 52.00 & 52.85 & 85.11 \\
    \textbf{A}: satnet\_data, \textbf{B}: multiple\_sol & 0.80 & 1.00 & 8.20 \\
    \textbf{A}: minimal\_17, \textbf{B}: big\_kaggle & 10.58 & 2.52 & 22.63 \\
    \textbf{A}: minimal\_17, \textbf{B}: multiple\_sol & 0.70 & 0.00 & 3.00 \\
    \textbf{A}: minimal\_17, \textbf{B}: satnet\_data & 18.90 & 6.10 & 36.60 \\
    \bottomrule
  \end{tabular}
  \caption{Results of cross-dataset experiments on Sudoku. DDReasoner-SL is the initial checkpoint from a supervised learning process using a limited amount of data and a few training epochs. Building upon this foundation model, DDReasoner-SL* and DDReasoner-RL are then separately obtained by applying SL and RL training with the entire training set, respectively, until convergence. All subsequent notations for DDReasoner-SL* and DDReasoner-RL in the latter part of this paper also carry this same meaning. The metric in the table represents DDReasoner's performance on the test set of Dataset \textbf{B}, after it was trained on the training set of Dataset \textbf{A}.}
  \label{cross_dataset}
\end{table*}

\subsubsection{Maze}
\begin{figure*}
  \centering
  \includegraphics[width=0.7\linewidth]{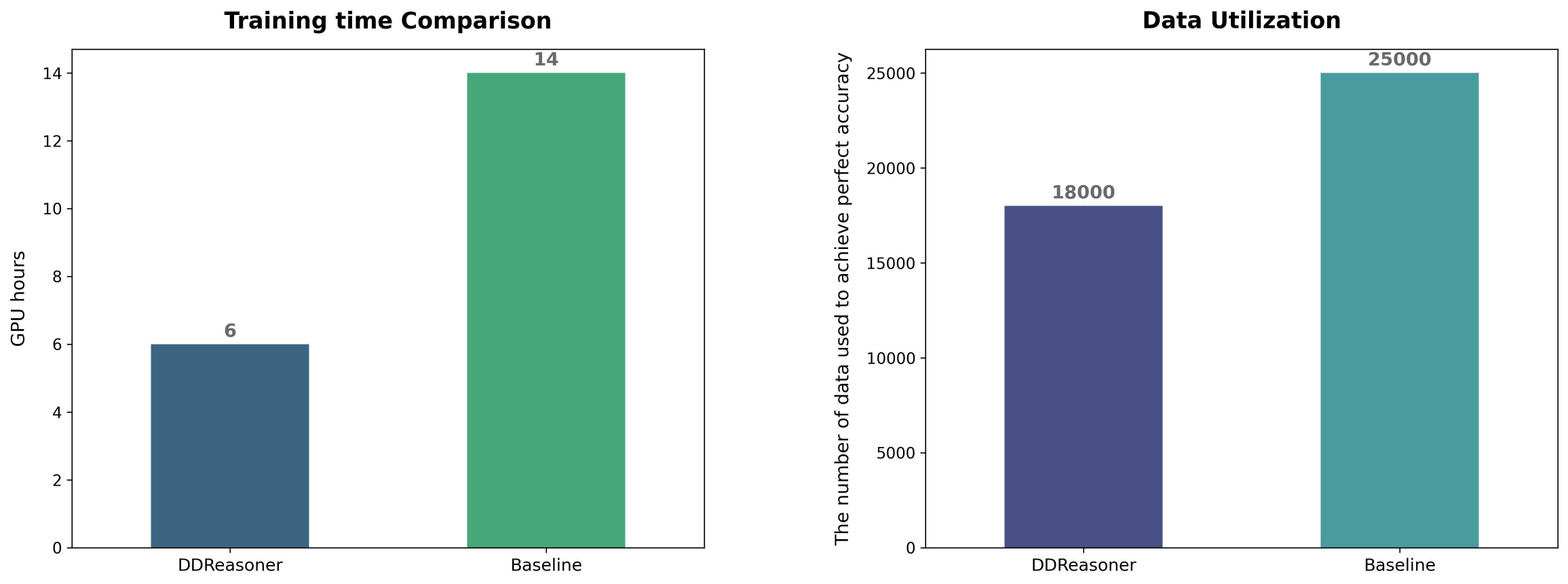}
  \caption{The efficiency comparison results on $5 \times 5$ Maze between DDReasoner and Baseline.}
  \label{result}
\end{figure*}

\paragraph{Efficiency Comparison}  We record the time of training our DDReasoner from scratch to reaching the perfect success rate, and compare it with the Baseline. In Figure \ref{result}, the training time for DDReasoner is 6 GPU hours, in which supervised learning requires 2 GPU hours while RL training requires 4 GPU hours.

\paragraph{Experiments across Multiple Grid Sizes}
To validate the impact of training with RL methods on the generalization ability of DDReasoner, we conducted the following experiments: DDReasoner was trained on mazes with a grid size of $M \times M$ and tested on mazes with a grid size of $N \times N$. Notably, to ensure data format uniformity when dealing with varying grid sizes, mazes of smaller grid sizes are randomly embedded at arbitrary locations within those of larger grid sizes. Hence, the test results from this experimental setup may exhibit a certain degree of randomness. The results of this experiment are presented in Table \ref{cross_size}.

\begin{table}
  \centering
  \begin{tabular}{llll}
    \toprule
    set  & DDReasoner-SL & DDReasoner-RL \\
    \midrule
    $M$: 20, $N$: 10 & $51.08\ {\pm\ 0.36}$  & $98.19\ {\pm\ 0.36}$ \\  
    $M$: 20, $N$: 5 & $39.43\ {\pm\ 0.73}$  & $99.03\ {\pm\ 0.03}$ \\  
    $M$: 5, $N$: 10 & $4.60$ & $33.55$ \\  
    $M$: 10, $N$: 20 & $12.60$ & $52.45$ \\ 
    \bottomrule
  \end{tabular}
  \caption{Results of experiments across multiple grid sizes on Maze. The metric in the table represents DDReasoner's performance on the test set of mazes with a grid size of $N \times N$, after the model was trained on the training set of mazes with a grid size of $M \times M$. When the grid size of the test set ($N$) is less than that of the training set ($M$), metrics will exhibit a certain degree of randomness; therefore, we conduct 10 experiments and present their statistical results, which include error bars.}
  \label{cross_size}
\end{table}

From the results presented above, it is apparent that the superiority of our reinforcement learning algorithm of fine-tuning DDReasoner only shows a less pronounced advantage in $5 \times 5$ mazes. Nevertheless, with escalating maze grid dimensions (and thereby increasing problem complexity), substantial performance gains become discernible, attributable to the combined efficacy of DDReasoner structure and the applied policy optimization algorithm in addressing such tasks. Furthermore, DDReasoner trained with our RL method exhibits significantly stronger generalization ability on mazes of varying complexities compared to the conventional SL method.

\subsubsection{Simple Path Prediction \& Preference Learning}
The ablation studies of the two tasks are shown in Table \ref{grid_result} and Table \ref{preference_result}. In table \ref{preference_result}, DDReasoner-SL and DDReasoner-SL* produce identical results. This indicates that continuing supervised learning on DDReasoner-SL leads to overfitting without yielding any improvement in metrics. Therefore, we retain the results from DDReasoner-SL for DDReasoner-SL*.

\begin{table}
  \centering
  \begin{tabular}{llll}
    \toprule
    model  & exact  & hamming & consistent \\
    \midrule
    DDReasoner-SL* &  67.70 & 94.28 &  74.84 \\
    DDReasoner-RL & 70.81 & 93.96 & 80.12 \\
    \bottomrule
  \end{tabular}
  \caption{Results of ablation studies on simple path prediction.}
  \label{grid_result}
\end{table}

\begin{table}
  \centering
  \begin{tabular}{llll}
    \toprule
    model  & exact  & hamming & consistent \\
    \midrule
    DDReasoner-SL*  & 21.02 & 78.15 &  99.80 \\
    DDReasoner-RL & 22.54 & 78.57 & 100.00\\
    \bottomrule
  \end{tabular}
  \caption{Results of ablation studies on preference learning.}
  \label{preference_result}
\end{table}

\subsubsection{Minimum-Cost Path Finding}
The ablation studies of the Minimum-Cost Path Finding task is shown in Table \ref{war_ablation}.

\begin{table}
  \centering
  \begin{tabular}{llllll}
  \toprule
    model  & &  k=12& k=18 &k=24 & k=30 \\
    \midrule
    \multirow{2}{*}{DDReasoner-SL*} & I & 88.90 & 82.50 & 76.10 & 71.90 \\
    & II & 98.60 & 97.80 & 95.60 & 92.50 \\
    \midrule
    \multirow{2}{*}{DDReasoner-RL} & I & 90.40 & 86.20 & 81.30 & 80.10 \\
    & II & 99.10 & 99.10 & 99.10 & 98.40 \\
    \hline
\end{tabular}
\caption{Results of ablation studies on the Minimum-Cost Path Finding. I and II denote the percentage of finding the shortest path and a connected path, respectively.}
  \label{war_ablation}
\end{table}

\section{C \ \ \ \ Cases}
\begin{figure*}
  \centering
  \includegraphics[width=0.8\linewidth]{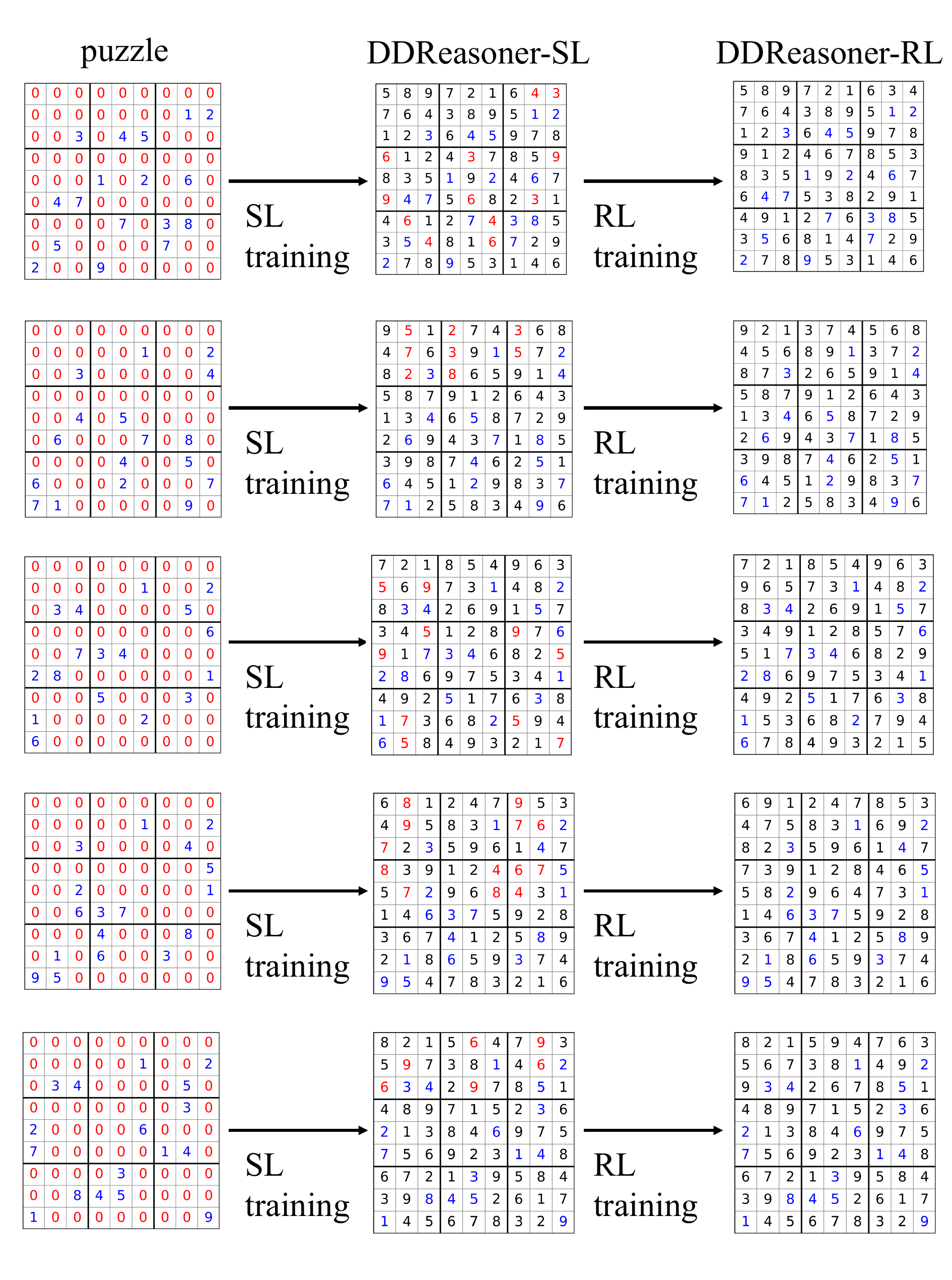}
  \caption{Examples of the effect of RL training on specific Sudoku cases. The blue numbers denote the hints already provided on the Sudoku board.}
  \label{sudokucase}
\end{figure*}

\begin{figure*}
  \centering
  \includegraphics[width=0.8\linewidth]{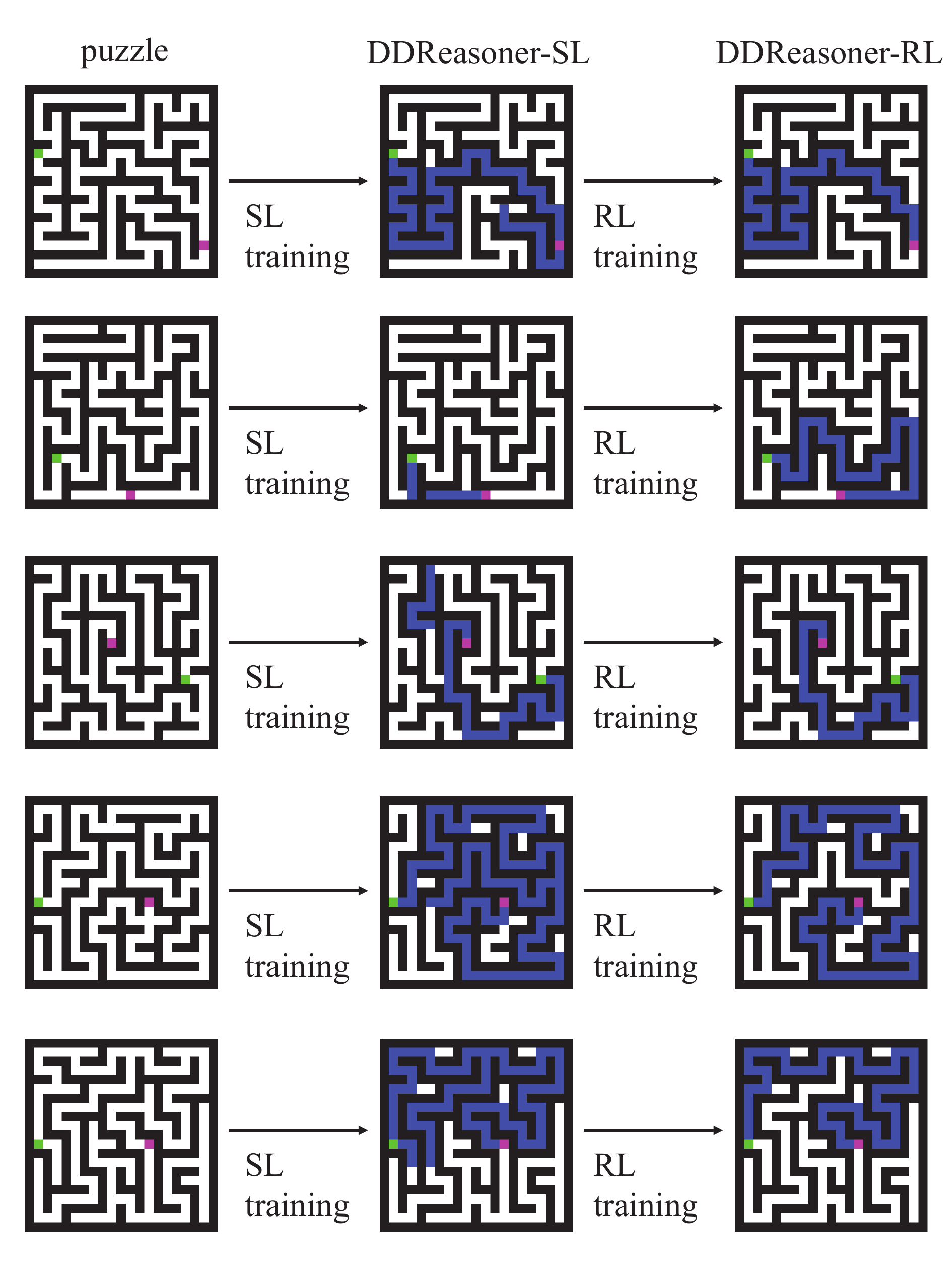}
  \caption{Examples of the effect of RL training on specific Maze cases.}
  \label{mazecase}
\end{figure*}

Figure \ref{sudokucase} and Figure \ref{mazecase} show some examples where problems, initially unsolved by supervised learning, are successfully addressed by our RL approach, which rectifies the inherent errors to produce valid solutions.

\section{D \ \ \ \ Discussions}
\label{discussion}
\paragraph{Why employ Reinforcement Learning?}While Supervised Learning (SL) provides an indispensable foundation, equipping DDReasoner with the basic generative priors for symbolic structures by mimicking known solutions, it is fundamentally insufficient for mastering the rigorous demands of complex symbolic reasoning tasks. The core challenge lies in the nature of these tasks: they often necessitate strict adherence to logical rules and constraints. Generated solutions with minor errors in a few positions, which lead to the violation of overall constraints, may exhibit a low loss during supervised learning. This suggests that supervised learning may fail to identify them as suboptimal outputs, whereas in practical applications, such outputs necessitate penalization. Therefore, our research introduced a novel application of reinforcement learning to fine-tune diffusion models for such complex reasoning tasks. Under this paradigm, only outputs that strictly satisfy global hard constraints receive positive rewards; conversely, any output violating these constraints, irrespective of the error's magnitude (e.g., a single incorrect position), is subject to penalization. Our experimental evaluations conducted across several classic symbolic reasoning benchmarks have demonstrated the crucial impact of RL training in steering neural outputs towards enhanced adherence to logical constraints.

\paragraph{When is it optimal to transition from SL to RL?}Factually, the transition point from the SL phase to the RL phase is of paramount importance and could critically determine the ultimate performance. If the SL phase is over-trained, the policy model's exploration capability will be reduced and thereby limiting the potential improvements achievable during the RL phase. Conversely, insufficient training in the SL phase can result in an inadequate initial reasoning ability, making it difficult to sample effective reasoning paths and thus lowering the upper bound of the model's potential that can be unlocked by RL. Therefore, optimizing data allocation and training strategies based on task characteristics is a practical issue requiring careful tuning.

\paragraph{Limitations}
\label{sec:limitations}
Although our method has demonstrated its effectiveness and efficiency on multiple symbolic reasoning benchmarks, the method still possesses some inherent limitations: (1) Insufficient Interpretability of Failure Cases: When DDReasoner-RL fails to solve a puzzle or provides an illogical solution, apart from knowing that it violated certain constraints, deeply understanding which specific step or pattern of judgment went awry during its multi-step denoising process remains a challenge; (2) Generalization Capability to Chanllenging Constraint Types: If the underlying logical rules of a new problem are exceptionally challenging to capture, and supervised learning is unable to adequately learn the complex underlying relationships, the subsequent RL phase may be rendered ineffective. In such circumstances, it might still be necessary to integrate a dedicated logical engine or a specific algorithmic implementation into the architecture, or require extensive targeted fine-tuning; (3) Preference for Deterministic and Unique Solutions: Our current evaluations are mainly focused on problems with deterministic (or a few) solutions. For symbolic problems that inherently have a large number of equivalent optimal solutions or require the generation of diverse compliant solutions, training objectives and sampling strategies might need adjustments to avoid mode collapse or learning only a narrow part of the solution distribution. 

These limitations do not negate the value of the our method but rather highlight key areas worthy of further research and refinement as we move towards more general and robust neuro-symbolic reasoning pipelines.

\paragraph{Opportunities for Future Advancement} Our methods of fine-tuning DDReasoner with RL techniques opens up several exciting avenues for future research:
\begin{enumerate}
    \item \textbf{Scaling to greater complexity and broader domains}: Future work can focus on extending DDReasoner-RL to tackle higher-dimensional problems, e.g., visual Sudoku and visual generation related to physical consistency. While scaling up to more complex spaces may necessitate the addition of an auxiliary neural network modules and logical constraint functions, the core architecture and training method proposed in this paper are generally applicable.

    \item \textbf{Enhancing interpretability and explainability}: While DDReasoner-RL demonstrates strong performance, further research into understanding its decision-making process during the denoising steps could provide valuable insights into how neural networks internalize and apply logical rules. Since our binary reward is given only at the end of the denoising process, a future direction could be to discretize intermediate steps of the denoising process and provide corresponding rewards. This approach may yield final outcomes that are more reliable and well-grounded.
    
    \item \textbf{Refining training dynamics and efficiency}: Further exploration into optimizing the interplay between the SL and RL stages, including the precise timing of the transition and the design of even more sophisticated reward or curriculum learning strategies for the RL phase, could lead to further gains in sample efficiency and overall performance.
\end{enumerate}

\paragraph{Summary}In conclusion, DDReasoner-RL provides a solid starting point and a rich space for imagination in leveraging RL methods to fine-tune diffusion models to solve symbolic reasoning problems. By overcoming existing challenges and exploring new research directions, such methods can hopefully be pushed towards broader and more complex application scenarios, like logistics, path planning and automated design, etc.

\end{document}